\documentclass{article}

\usepackage[preprint]{neurips_2026}

\usepackage[utf8]{inputenc} 
\usepackage[T1]{fontenc}    
\usepackage{url}            
\usepackage{booktabs}       
\usepackage{amsfonts}       
\usepackage{nicefrac}       
\usepackage{microtype}      
\usepackage{color}
\usepackage[table]{xcolor}
\usepackage{multirow}
\usepackage{graphicx}
\usepackage{amsmath}
\usepackage{makecell}
\usepackage{pifont}
\usepackage{subcaption}
\usepackage{algorithm}
\usepackage{algpseudocode}
\usepackage{wrapfig}

\definecolor{citecolor}{HTML}{0071BC}
\definecolor{linkcolor}{HTML}{ED1C24}
\usepackage[pagebackref=false, 
            breaklinks=true,
            letterpaper=true,
            colorlinks,
            citecolor=citecolor, 
            linkcolor=linkcolor, 
            bookmarks=false]{hyperref}

\definecolor{lightgray}{gray}{0.7}
\definecolor{mygreen}{rgb}{0, 0.69, 0.31}
\definecolor{citecolor}{HTML}{0071BC}
\definecolor{linkcolor}{HTML}{ED1C24}
\definecolor{darkergreen}{RGB}{21, 152, 56}
\definecolor{red2}{RGB}{252, 54, 65}
\colorlet{defaultcolor}{purple!10}

\newcommand{\yes}{\textcolor{darkergreen}{\ding{52}}}
\newcommand{\no}{\textcolor{red2}{\ding{56}}}

\newcommand{\ours}{NOPD}

\newlength\savewidth
\newcommand\shline{
  \noalign{
    \global\savewidth\arrayrulewidth
    \global\arrayrulewidth 1pt
  }
  \hline
  \noalign{
    \global\arrayrulewidth\savewidth
  }
}

\title{Self-Boosting Vision-Language Models with \\Noisy Student On-Policy Self-Distillation}

\author{%
Shuai Wang$^1$ \quad
Daoan Zhang$^2$ \quad
Zhe Tang$^3$ \quad 
Hao Cheng$^4$ \quad 
Jiaheng Wei$^1$ \\
$^1$ The Hong Kong University of Science and Technology (Guangzhou) \quad $^2$ University of Rochester \\
$^3$ Zhejiang University of Technology \quad $^4$ Hong Kong Baptist University \\
}

\begin{document}
\maketitle
\begin{abstract}
Post-training enables vision-language models (VLMs) to understand human instructions and perform various downstream tasks. Current  post-training methods usually rely on human-annotated data, distillation from external models, reinforcement learning with human feedback, or verifiable answers. This limits their ability to improve without external supervision. To tackle this, we propose NOPD (Noisy Student On-Policy Self-Distillation), a simple yet effective self-distillation approach that improves VLMs without any external models or ground-truth answers. Our key insight is that prediction discrepancies between clean and corrupted inputs naturally induce a self-supervision signal. In NOPD, the model learns from corrupted inputs while using its own predictions under clean inputs as token-level supervision. We show the effectiveness of NOPD on five visual reasoning tasks; it can match and even outperform reinforcement learning approaches or distillation from external models. Notably, when trained with 2.1K samples from Geometry3K, NOPD improves Qwen2.5-VL-7B by 20 points on its validation set. It also shows generalization on out-of-distribution test sets and achieves 7.4 point gains on MathVista. Furthermore, we demonstrate that NOPD is a general approach to enhance VLMs, achieving improvements across three models on 12 benchmarks.
\end{abstract}
\section{Introduction}
Vision-language models (VLMs) have made remarkable progress in visual question answering (VQA), visual reasoning and chart understanding~\citep{qwen2p5-vl,glm45v,llava}. The advancement is primarily achieved in the post-training stage, which may involve supervised fine-tuning (SFT)~\citep{llava}, reinforcement learning with human feedback~\citep{instructgpt} or verifiable rewards (RLVR)~\citep{guo2025deepseekr1,olmo2,shao2024deepseekmath}, and knowledge distillation~\citep{agarwal2024onpolicy,gu2024minillm}.

Although recent research has advanced the foundational capabilities of VLMs, typical post-training approaches exhibit inherent limitations. SFT usually needs human annotation or distillation from frontier models. RLVR typically requires verifiable questions and only provides a sparse reward signal for the whole response. Off-policy and on-policy knowledge distillation provide stronger and denser token-level supervision from an external model (usually called \textit{teacher}). 

Given these limitations and the strong capabilities of modern VLMs, we ask: can a VLM improve itself without any ground-truth answers or supervision from external models, i.e., ``self-boosting''? Recently, some self-distillation approaches have been proposed to improve models without external models by using environment feedback~\citep{reinforcement_sd}, or using answers as additional context~\citep{self_distillation_reasoner}. They are mainly designed for language models and still rely on external information. We aim to improve VLMs without any external supervision or feedback.

\begin{figure}[t]
  \centering
  \begin{minipage}{\textwidth}
    \centering
    \includegraphics[width=\textwidth]{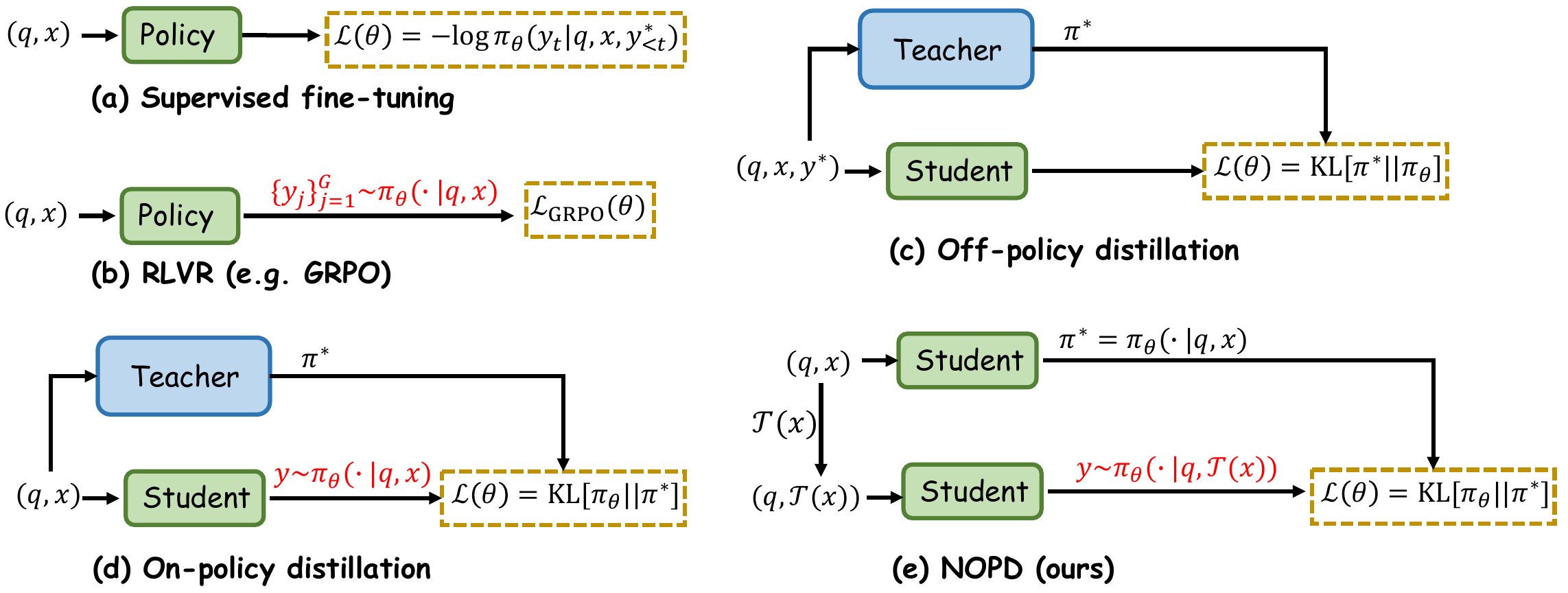}
  \end{minipage}
  \vfill
  \vspace{0.6cm}
  \begin{minipage}{\textwidth}
    \centering
      \begin{tabular}{lccccc}
        & No GT & On-policy & No external  & Token-level  & Low sampling\\
        & needed    &     data      & models  & learning signal        & cost \\
        \shline
        SFT & \no & \no & \yes & \yes & \yes \\
        RLVR & \no & \yes & \yes & \no & \no \\
        KD & \yes & \no & \no & \yes & \yes \\
        OPD & \yes & \yes & \no & \yes & \yes \\
        \ours{} (ours) & \yes& \yes & \yes & \yes & \yes \\
      \end{tabular}
  \end{minipage}
  \caption{\textbf{Comparison of different post-training approaches for VLMs.} \textbf{Top}: demonstration of different approaches. $q$ denotes text query and $x$ denotes image. We omit the subscripts of loss functions except GRPO for simplicity. We highlight all sampling operations with \textcolor{red}{red}, e.g., \textcolor{red}{$y\sim \pi_{\theta}(\cdot|q,x)$}. $\pi_{\theta}$ denotes the policy model in (a) and (b), and the student model for distillation in (c), (d), and (e). $\pi^{*}$ denotes the teacher for distillation methods. In (e), $\mathcal{T}$ denotes a pre-defined transformation. (e) gives an overview of our method \ours. Different from typical on-policy distillation, \ours{} shares the same parameters between student and teacher. The teacher supervision signal comes from the model under raw inputs $(q,x)$. For the student model, inputs are corrupted images $\mathcal{T}(x)$ and text query $q$. \textbf{Bottom}: summary of comparison. GT denotes answers for given input questions, and may include detailed chain-of-thought~\citep{cot} traces.}
  \label{fig:compare_approach}
\vspace{-1em}
\end{figure}

We show that supervision can be self-induced by exploiting prediction discrepancies under asymmetric input distributions. Specifically, when a model processes clean and corrupted inputs, the resulting difference in output distributions naturally provides a dense learning signal—without requiring labels, rewards, or external teachers. We instantiate this principle and propose noisy student on-policy self-distillation (\ours), a self-improvement approach for VLMs. In \ours, a single model plays both teacher and student roles. The teacher works with clean inputs, while the student is trained using corrupted ones. The predictions under clean inputs serve as token-level supervision for the noisy inputs. Compared with previous methods, \ours{} obviates the need for high-quality responses and external strong models as shown in Figure~\ref{fig:compare_approach}.

We conduct extensive experiments to validate the effectiveness of \ours. We first conduct preliminary experiments to compare \ours{} with different post-training methods on five visual reasoning benchmarks. Extensive experiments demonstrate that \ours{} consistently improves VLMs across multiple benchmarks, achieving performance comparable to or exceeding methods that rely on external supervision. Notably, \ours{} also leads to implicit reasoning compression, reducing output length while improving accuracy. These results suggest that effective post-training can be achieved without any external supervision. Finally, we validate whether \ours{} generalizes across different models and tasks beyond reasoning.
\section{Preliminaries}
In this section, we briefly review existing post-training paradigms.
\paragraph{Notation.}
We use $q$ to denote a text query, and $x$ denotes an image from the training set $\mathcal{S}$. A vision-language model parameterized by $\theta$ is defined as $\pi_{\theta}$. Given an input pair $\left( q, x \right)$, its likelihood under $\pi_{\theta}$ is $\pi_\theta(y \mid q, x) = \prod\limits_{t=1}^{\lvert y \rvert} \pi_\theta\left(y_t \mid q, x, y_{<t}\right)$, where $\lvert y \rvert$ denotes the number of tokens in $y$. $y^{*}$ denotes the ground-truth response for input pair $\left(q,x\right)$, which is typically obtained from human labeling or distillation from frontier models.

\paragraph{Supervised fine-tuning (SFT).}
SFT is widely used in post-training of modern foundation models that minimizes the following negative log-likelihood loss
\begin{equation}
    \mathcal{L}_\text{SFT}\left(\theta\right) = -\sum_{t=1}^{\lvert y^{*} \rvert} \log \pi_{\theta}\left(y_t \mid q,x,y^{*}_{<t}\right).
\end{equation}
Note that if $y^{*}$ is generated by other models rather than human annotation, it is also called sequence-level knowledge distillation~\citep{kim2016sequencekd}. SFT provides token-level supervision to guide the learning with the help of $y^*$.

\paragraph{Reinforcement learning with verifiable rewards (RLVR).}
RLVR~\citep{olmo2} is an on-policy reinforcement learning approach used for specific domains where questions have verifiable answers, such as math and code. The representative algorithm is GRPO~\citep{shao2024deepseekmath}, which computes advantages in a relative group instead of using additional critic models. The GRPO objective is formulated as
\begin{align}
\label{equ:grpo}
\mathcal{L}_\textrm{GRPO}\left(\theta\right)
= \mathbb{E}_{ (q,x) \sim \mathcal{S}, \, \{y_i\}_{i=1}^G \sim \pi_{\theta_\textrm{old}}\left(\cdot \mid q,x\right) }
\Bigg[
\frac{1}{G} \sum_{i=1}^{G} \frac{1}{\lvert y_i\rvert} 
\sum_{t=1}^{\lvert y_i\rvert} 
\min \Big(
    w_{i,t}\left(\theta\right) \widehat{A}_{i,t}, \notag \\
    \mathrm{clip}\big( w_{i,t}\left(\theta\right), 1 - \varepsilon, 1 + \varepsilon \big) \widehat{A}_{i,t} 
\Big)
\Bigg],
\end{align}
where $G$ is the number of generated responses for each input pair $\left(q,x\right)$ (i.e., the group size) and $\varepsilon$ is a hyperparameter to control the clipping range for stabilizing training. The importance ratio $w_{i,t}(\theta)$ and advantage $\widehat{A}_{i,t}$ of token $y_{i,t}$ are:
\begin{align}
    w_{i,t}\left(\theta\right)=\frac{ \pi_{\theta} \left(y_{i,t} \mid q,x, y_{i,<t}\right) }{ \pi_{\theta_\text{old}} \left(y_{i,t} \mid q,x,y_{i,<t}\right)},\quad
    \widehat{A}_{i,t} = \widehat{A}_{i} = \frac{r\left(q,x, y_i\right) - \mathrm{mean} \left( \{ r\left(q,x, y_i\right) \}_{i=1}^G \right) }{ \mathrm{std} \left( \{ r\left(q,x, y_i)\right \}_{i=1}^G \right) },
\end{align}
respectively, where all the tokens in $y_i$ share the same advantage as $\widehat{A}_{i}$. $r\left(q,x,y_i\right)$ denotes the reward for given inputs $(q,x)$ and responses $y$ generated by $\pi_{\theta}$. RLVR benefits from on-policy learning while suffering from sparse supervision signals and high sampling cost to rollout during training.

\paragraph{Off-policy knowledge distillation.} 
The goal of knowledge distillation (KD)~\citep{hinton2015kd} is to transfer knowledge from the strong model (usually named \textit{teacher}) to the weak model (usually named \textit{student}). Standard KD can be formulated as: 
\begin{equation}
    \mathcal{L}_{\textrm{KD}}\left(\theta\right) = \mathbb{E}_{\left(q,x\right) \sim \mathcal{S},\, y^{*} } 
       \left[ \mathcal{D}_{\textrm{KL}}\left(\pi^{*}\left(y^{*}\mid q,x\right)\|\pi_{\theta}\left(y^{*}\mid q,x\right)  \right)\right],
\end{equation}
where $\pi^{*}$ denotes the teacher model and $\mathcal{D}_{\textrm{KL}}(P\|Q)$ denotes the Kullback--Leibler (KL) divergence.

\paragraph{On-policy distillation (OPD).}
OPD~\citep{agarwal2024onpolicy,gu2024minillm} trains the student model with self-generated sequences and learns feedback from the teacher on these sequences. Specifically, OPD minimizes an objective of the form:
\begin{equation}
    \mathcal{L}_{\text{OPD}}\left(\theta\right) = \mathbb{E}_{\left(q,x\right) \sim \mathcal{S},\, y\sim \pi_{\theta}\left( \cdot \mid q,x\right) } 
       \left[ \mathcal{D}_{\textrm{KL}}\left(\pi_{\theta}\left(y\mid q,x\right) \| \pi^{*}\left(y\mid q,x\right)\right)\right].
\end{equation}
Off-/on-policy distillation provides dense token-level supervision without requiring human annotations, but typically relies on a stronger teacher model. When the teacher is derived from the same model as the student (e.g., sharing parameters) rather than an external model, we refer to this setting as on-policy self-distillation~\citep{self_distillation_reasoner,reinforcement_sd}. In this case, privileged information can be transferred through differences in inputs~\citep{jmlr_vapnik_kd}.
\begin{algorithm}[t]
\caption{Training Procedure of \ours}
\label{alg:training}
\begin{algorithmic}[1]
\Require Training set $\mathcal{S}=\{(q_i,x_i)\}_{i=1}^{|S|}$; vision-language model $\pi_{\theta}$; batch size $B$; image transformation $\mathcal{T}$; max iterations $T_{\max}$
\Ensure Trained model $\pi_{\theta}$
\State Initialize $t \leftarrow 0$
\While{$t < T_{\max}$ and not converged}
    \State Sample a mini-batch $\{(q_i,x_i)\}_{i=1}^{B}$ from $\mathcal{S}$
    \State Sample on-policy response $y_i \sim \pi_{\theta}\left(\cdot \mid q_i,\mathcal{T}\left(x_i\right)\right), \quad i=1,\dots,B$
    \State Compute the loss as Eq.~\eqref{eq:noisy_opd}
    \[
    \mathcal{L}\left(\theta\right) = \mathbb{E}_{\left(q,x\right) \sim \mathcal{S},\, y\sim \pi_{\theta}\left( \cdot | q,\mathcal{T}(x)\right)} 
       \left[ \mathcal{D}_{\textrm{KL}}\left(\pi_{\theta}\left(y\mid q,\mathcal{T}\left(x\right)\right) \| \pi_{\theta}\left(y\mid q,x\right)\right)\right]
    \]
    \State Update model parameters $\theta$ using $\nabla_{\theta}\mathcal{L}$ (c.f. Eq.~\eqref{eq:grad_nopd})
    \State $t \leftarrow t + 1$
\EndWhile
\State \Return $\pi_{\theta}$
\end{algorithmic}
\end{algorithm}
\begin{figure}[t]
    \centering
    \begin{minipage}[b]{0.24\textwidth}
        \centering
        \includegraphics[width=\textwidth]{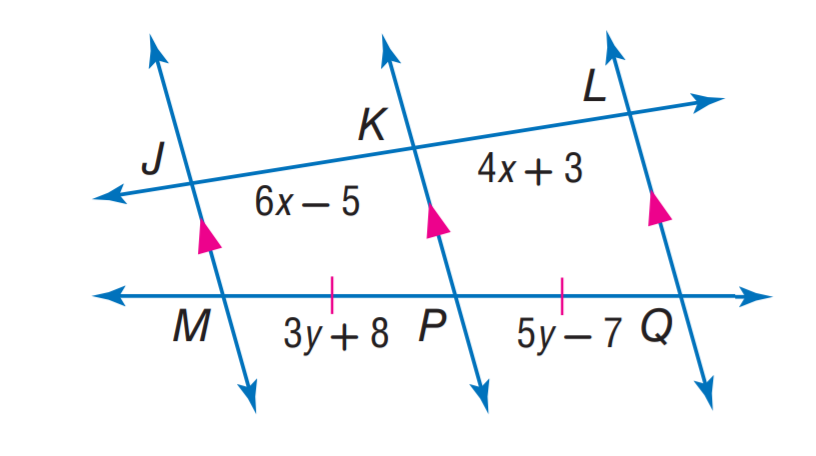}
    \end{minipage}
    \hfill
    \begin{minipage}[b]{0.24\textwidth}
        \centering
        \includegraphics[width=\textwidth]{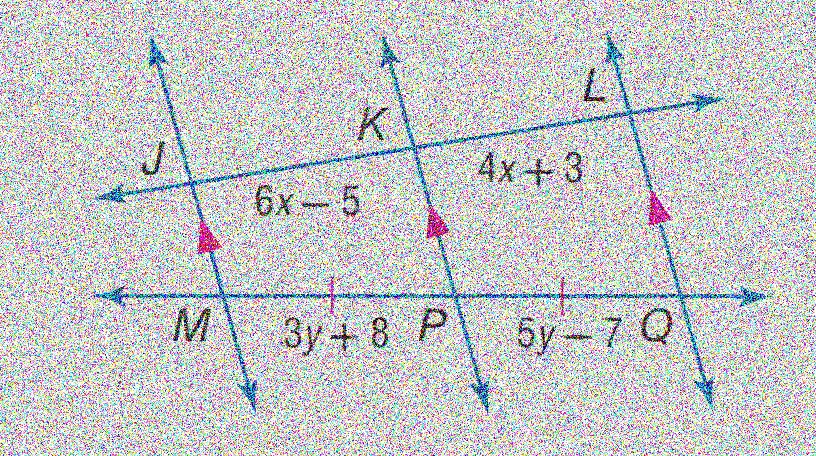}
    \end{minipage}
    \hfill
    \begin{minipage}[b]{0.24\textwidth}
        \centering
        \includegraphics[width=\textwidth]{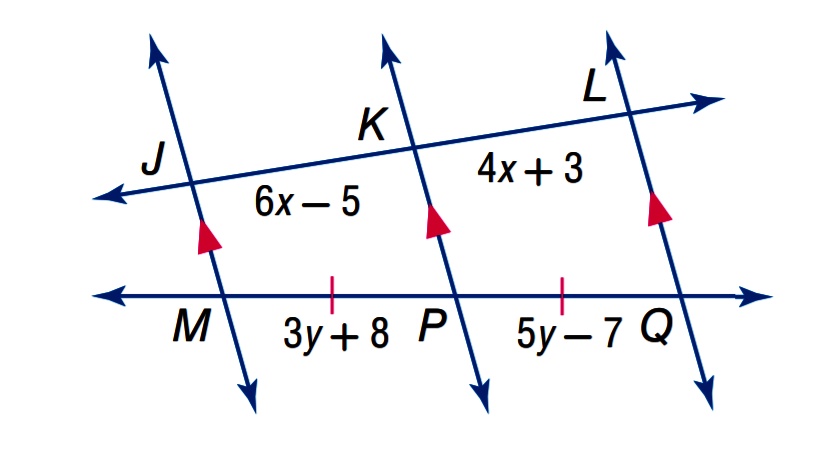}
    \end{minipage}
    \hfill
    \begin{minipage}[b]{0.24\textwidth}
        \centering
        \includegraphics[width=\textwidth]{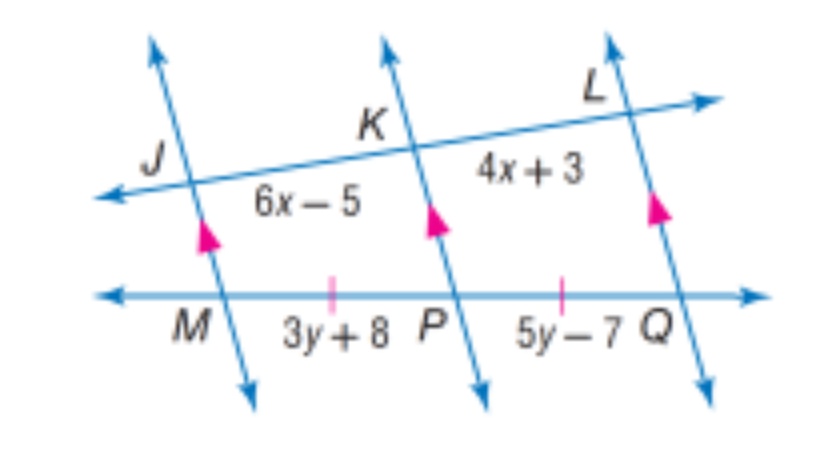}
    \end{minipage}
    \caption{Examples of different distorted images. The types of image distortions (from left to right): original images, add noise, gamma transformation, downsample. We resize the downsampled image for clarity. The image is from the training set of Geometry3K.}
    \label{fig:example_corrupt_images}
\vspace{-1em}
\end{figure}
\section{Method}
We propose \ours, short for \textbf{n}oisy student \textbf{o}n-\textbf{p}olicy self-\textbf{d}istillation, a self-training approach that enables VLMs to improve themselves without any ground-truth answers or external models. Algorithm~\ref{alg:training} provides an overview of \ours. The inputs to the student in the algorithm are corrupted images $\mathcal{T}(x)$ and the corresponding text query $q$. We first generate an on-policy response $y$ under corrupted images $y \sim \pi_{\theta}\left(\mathcal{T}\left(x\right), q\right)$, where $\mathcal{T}$ is a pre-defined transformation for images. After that, we use the outputs of the same model $\pi_{\theta}$ under clean inputs $(q,x)$ as the ``teacher'' output to provide dense supervision signals. Finally, our objective is to minimize the divergence between the student model and the teacher model as follows
\begin{equation}
    \mathcal{L}_{\textrm{NOPD}}\left(\theta\right) = \mathbb{E}_{\left(q,x\right) \sim \mathcal{S},\, y\sim \pi_{\theta}\left( \cdot \mid q,\mathcal{T}\left(x\right)\right)} 
       \left[ \mathcal{D}_{\textrm{KL}}\left(\pi_{\theta}\left(y\mid q,\mathcal{T}\left(x\right)\right) \| \underbrace{\pi_{\theta}\left(y\mid q,x\right)}_{\textcolor{blue}{\texttt{sg}}}\right)\right],
\label{eq:noisy_opd}
\end{equation}
where \textcolor{blue}{\texttt{sg}} denotes stop gradient. Note that the KL divergence is not symmetric, which means that $\mathcal{D}_{\textrm{KL}}\left(P \| Q\right) \neq \mathcal{D}_{\textrm{KL}}\left(Q\| P\right)$, where $P=\pi_{\theta}\left(y\mid q,\mathcal{T}\left(x\right)\right)$ and $Q=\pi_{\theta}\left(y\mid q,x\right)$. The KL divergence in Eq.~\eqref{eq:noisy_opd} is called reverse KL. We consider two other commonly used variants: forward KL $\mathcal{D}_{\textrm{KL}}\left( Q \|P \right)$ and generalized Jensen--Shannon divergence (JSD), which is defined as
\begin{equation}
    \mathcal{D}_{\textrm{JSD}}\left(P \| Q\right)\left(\beta\right) = \beta D_{\mathrm{KL}}\left(P \,\|\, \beta P + \left(1-\beta\right) Q \right) 
+ \left(1-\beta\right) D_{\mathrm{KL}}\left(Q \,|\, \beta P + \left(1-\beta\right) Q \right),
\end{equation}
where $\beta \in (0,1)$ is the weighting coefficient. The gradients of JSD are similar to forward KL when $\beta\rightarrow 0$ and reverse KL when $\beta\rightarrow 1$, respectively. We verify the effectiveness of different divergences in Sec.~\ref{sec:exp_analysis}. We stop the gradient from the ``teacher'' signal, i.e., $\pi_{\theta}\left(y\mid q,x\right)$, to prevent model collapse\footnote{If we do not stop the gradient of $\pi_{\theta}\left(y\mid q,x\right)$, $\pi_{\theta}$ will generate the same output whatever the input to make the loss zero.} and do not backpropagate through the student's sampling distribution $\pi_{\theta}(\cdot \mid q,\mathcal{T}(x))$, which makes the training process more stable and efficient following~\citep{agarwal2024onpolicy}.

Distinguished from other post-training approaches, our method \ours{} is a self-training approach to improve VLMs without any external supervision (models, verifiable answers, and reasoning traces). In addition, \ours{} also has low sample cost, is more data efficient, and could reduce output length with performance increase; see Sec.~\ref{sec:exp_analysis} for verification. We compare our method with other post-training approaches in Figure~\ref{fig:compare_approach} and provide more discussion in Sec.~\ref{sec:related_work}.

Our key contribution lies in introducing asymmetric input distributions by adding permutation $\mathcal{T}$ for the student vision inputs  to enable self-induced supervision, which makes our method different from other self-distillation methods~\citep{self_distillation_reasoner,reinforcement_sd}. The motivation behind this is that we aim to improve student models by exposing them to difficult environments in terms of noise during learning. Learning from asymmetric input distributions to enable self-induced supervision also improves generalization to out-of-distribution domains.

\paragraph{Gradient Analysis.}
We can derive the \ours{} gradient as follows (see Sec.~\ref{sec:appendix_results_analysis} for details):
\begin{align}
\small
\nabla_{\theta} \mathcal{L}_{\text{NOPD}}(\theta) =
&\;\mathbb{E}_{(q,x)\sim \mathcal{S}, \; y \sim \pi_\theta(\cdot \mid q,\mathcal{T}(x))}
\Bigg[\sum_{t=1}^{\vert y\rvert} 
\mathbb{E}_{\hat{y}_t \sim \pi_\theta(\cdot \mid \mathcal{T}(x), q, y_{<t})} \nonumber \\
&\;\quad 
\Bigg[\log \frac{\pi_\theta(\hat{y}_t \mid q, \mathcal{T}(x))}
         {\pi_\theta(\hat{y}_t \mid q, x)} 
\cdot \nabla_\theta \log \pi_\theta(\hat{y}_t \mid q, \mathcal{T}\left(x\right), y_{<t})
\Bigg] 
\Bigg],
\label{eq:grad_nopd}
\end{align}
where $\hat{y}_t \in \Delta^{\lvert V \rvert-1}$ is $\lvert V \rvert$-simplex and denotes the model's predicted probability distribution over the vocabulary at decoding step $t$, and $\lvert V \rvert$ is the vocabulary size.

The \ours{} gradient can be interpreted as a contrastive policy gradient. The inner term, \(\nabla_\theta \log \pi_\theta(\hat{y}_t \mid q, \mathcal{T}(x), y_{<t})\), is the standard policy gradient that propagates the sampled token \(\hat{y}_t\) back to the model parameters. The log-ratio term $\log \dfrac{\pi_\theta(\hat{y}_t \mid q, \mathcal{T}(x))}{\pi_\theta(\hat{y}_t \mid q, x)}$ acts as a contrastive reward, amplifying updates for tokens whose predicted distributions differ significantly between the transformed input $\left(q,\mathcal{T}\left(x\right)\right)$ and the original input $\left(q,x\right)$. 
\vspace{-.5em}
\paragraph{The choice of image permutation.}
For image transformation $\mathcal{T}$, we consider three ways:

$\bullet$ Additive Gaussian noise. An augmented image $x'$ is generated by adding Gaussian noise to the input image:
$x' = \text{clip}(x + \epsilon, 0, 1)$, where $\epsilon \sim \mathcal{N}(0, \sigma^2)$.
The noise standard deviation $\sigma$ is randomly sampled from $\{0.08, 0.12, 0.18, 0.26, 0.38\}$, following~\citep{imagenet-c}.

$\bullet$ Lower resolution. We randomly downsample the input image $x$ by a factor of $2$ or $4$, resulting in a lower-resolution image. This process reduces image fidelity and decreases the number of vision tokens, thereby improving training efficiency.

$\bullet$ Gamma transformation. We apply gamma transformation to adjust the brightness and contrast of given images. To be specific, the augmented image is $x^{\prime}=x^{\gamma}$ and we sample $\gamma \sim U[0.5,2]$.

We provide examples of distorted images in Figure~\ref{fig:example_corrupt_images}. Note that we choose hyperparameters for image transformation $\mathcal{T}$ directly widely used in computer vision tasks and do not tune hyperparameters due to the limitation of compute resources. During training, we randomly select one transformation and apply it to get corrupted image $x$.
\begin{table}[t]
\centering
\caption{\textbf{Comparison of different approaches.} We highlight the \textbf{best} result and \textcolor{lightgray}{performance degradation} compared with the base model.}
    \begin{tabular}{lcccccc}
    \toprule
     & \multicolumn{2}{c}{\textbf{Geometry3K}}  & \textbf{MathVista} & \textbf{MathVision}  & \textbf{MMMU}  & \textbf{MMMU-Pro} \\
     & test & val & testmini & test & val & overall \\
    \midrule
    Qwen3.5-2B & 73.2 & 65.3 & 76.1 & 44.3 & 64.2 & 50.3 \\
    SFT        & 73.2 & 66.0 & \textcolor{lightgray}{74.8} & \textcolor{lightgray}{42.8} & \textcolor{lightgray}{63.8} & \textcolor{lightgray}{49.4}\\
    KD         & 74.0 & 65.7 & \textcolor{lightgray}{75.4} & \textcolor{lightgray}{43.3} & 64.9 & 50.8 \\
    RLVR       & \textbf{78.4} & 68.2 & 79.3 & \textbf{48.1} & 67.0 & 53.5\\
    OPD        & 77.2 & 67.8 & 78.7 & 46.6 & \textbf{68.1} & \textbf{55.3} \\    
    \ours{} (ours)    & 78.0 & \textbf{69.1} & \textbf{80.4} & 47.2 & 67.6 & 54.7 \\
    \midrule
    Qwen2.5-VL-7B & 35.4 & 33.7 & 68.4 & 25.1 & 58.6 & 38.3 \\
    SFT           & 38.8 & 37.0 & 70.2 & 26.2 & 59.1 & 38.3 \\
    KD            & 39.1 & 37.3 & 70.2 & 25.8 & 59.1 & \textcolor{lightgray}{38.1} \\    
    RLVR          & 52.1 & 51.0 & 72.3 & 28.1 & 62.8 & 41.8 \\
    OPD           & 54.8 & 52.0 & 73.1 & 28.1 & \textbf{63.6} & 42.0 \\    
    \ours{} (ours)         & \textbf{55.2} & \textbf{53.7} & \textbf{75.8} & \textbf{30.9}  & 63.2 & \textbf{43.3} \\
    \bottomrule
    \end{tabular}
\label{tab:simple_exp}
\vspace{-1em}
\end{table}
\section{Experiments}
We conduct extensive experiments to answer the following questions:

$\bullet$ How does \ours{} compare to other post-training approaches? (Sec.~\ref{sec:exp_pre_reason})

$\bullet$ How do the design choices affect performance? (Sec.~\ref{sec:exp_analysis})

$\bullet$ Is \ours{} a general approach to improve VLMs itself in other domains? (Sec.~\ref{sec:exp_general})
\label{sec:exp}
\subsection{A Preliminary Experiment on Reasoning Tasks}
\label{sec:exp_pre_reason}
In this section, we conduct a preliminary experiment to verify the effectiveness of our method and compare it with other post-training approaches including SFT, off-policy KD, RLVR and OPD.
\begin{table}[t]
\centering
\setlength{\tabcolsep}{5pt}
\caption{\textbf{Ablation studies} of key design choices. We train Qwen3.5-2B with the training set of Geometry3K and report accuracy on MathVista. Default settings are \colorbox{defaultcolor}{marked}.}
\begin{subtable}[t]{0.23\textwidth}
\vspace{0pt}
\centering
\begin{tabular}{lc}
Loss & Acc. \\
\shline
Forward KL & 78.1 \\
\rowcolor{defaultcolor}
Reverse KL &  80.4\\
JSD ($\beta=0.5$)  & 79.8 \\
\end{tabular}
\caption{Loss type}
\label{tab:ablation_loss}
\end{subtable}
\hfill
\begin{subtable}[t]{0.23\textwidth}
\vspace{0pt}
\centering
\begin{tabular}{lc}
 & Acc. \\
\shline
\rowcolor{defaultcolor}
Full vocab    &  80.4  \\
Sampled token &  79.1  \\
~ & ~ \\
\end{tabular}
\caption{Computation strategy}
\label{tab:ablation_compute}
\end{subtable}
\hfill
\begin{subtable}[t]{0.23\textwidth}
\vspace{0pt}
\centering
\begin{tabular}{lc}
 & Acc. \\
\shline
Fix  & 76.9\\
EMA  & 79.1\\
\rowcolor{defaultcolor}
Sync &  80.4\\
\end{tabular}
\caption{Teacher update}
\label{tab:ablation_teacher_update}
\end{subtable}
\hfill
\begin{subtable}[t]{0.23\textwidth}
\vspace{0pt}
\centering
\begin{tabular}{lc}
Gen length & Acc. \\
\shline
1024 &  79.4\\
\rowcolor{defaultcolor}
2048 &  80.4\\
4096 &  80.5\\
\end{tabular}
\caption{Generation Length}
\label{tab:ablation_gen_len}
\end{subtable}
\vspace{-1em}
\label{tab:ablation}
\end{table}
\paragraph{Models.}
We use Qwen2.5-VL-7B(-Instruct)~\citep{qwen2p5-vl} and Qwen3.5-2B\footnote{\url{https://huggingface.co/Qwen/Qwen3.5-2B}} as the base models. Note that Qwen3.5-2B is a hybrid model that supports thinking and no-thinking modes. To control for variables, we conduct experiments exclusively in the thinking mode.

\paragraph{Training details \& baselines.}
We use the Geometry3K training set~\citep{geo3k}, which contains 2.1K image-text pairs. For the SFT baseline, we use responses distilled from Qwen3.5-27B/Qwen2.5-VL-72B-Instruct for Qwen3.5-2B/Qwen2.5-VL-7B. For a fair comparison, we keep the same teacher model for off-policy KD and on-policy distillation experiments. For RLVR, we use GRPO~\citep{shao2024deepseekmath} as the standard implementation. We also experiment with other variants, such as DAPO~\citep{dapo} and GSPO~\citep{gspo}, but observe no significant improvement. Training hyperparameters are included in Table~\ref{tab:training_hp}.

\paragraph{Evaluation.}
We first evaluate on the test and val split of Geometry3K~\citep{geo3k} as the in-distribution test set. In addition, we add MathVista~\citep{lu2024mathvista}, MathVision~\citep{mathvision}, MMMU~\citep{mmmu} and MMMU-Pro~\citep{mmmu-pro} as out-of-distribution test sets. For Qwen2.5-VL-7B~\citep{qwen2p5-vl}, we use greedy decoding with CoT~\citep{cot} and report pass@1 accuracy. For Qwen3.5-2B, we follow the Qwen official guide to use random sampling and report avg@16 for stable performance. More details are included in Sec.~\ref{sec:appendix_eval_details}.
\paragraph{Results.}
Table~\ref{tab:simple_exp} reports results on five benchmarks. First, \ours{} consistently improves the base model across \textbf{all} benchmarks. Specifically, \ours{} improves 4.4 points on MMMU-Pro and 4.8 points on the test set of Geometry3K with Qwen3.5-2B. In addition, \ours{} achieves a 20-point improvement on the val split of Geometry3K and 7.4 points on MathVista with Qwen2.5-VL-7B. Trained only with the Geometry3K training set, \ours{} also improves performance on out-of-distribution test sets, which demonstrates strong out-of-domain generalization. 

Furthermore, \ours{} can match or even outperform RLVR and OPD on multiple benchmarks. We also find that performance degradation often occurs with SFT and off-policy KD, which indicates that they struggle to improve open-weight models even with the help of teacher models or external labels. RLVR and OPD consistently improve all benchmarks for both models, but they require external ground-truth answers and models. The results show that \ours{} is a self-training method that improves VLMs without any external signals (ground-truth answers or external models) and the effectiveness of \ours{} matches RLVR and OPD.

\subsection{Analysis}
\label{sec:exp_analysis}
\paragraph{Design choices.}
We conduct extensive experiments to study key design choices in our method and present the results in Table~\ref{tab:ablation}. First, a key design choice is the loss function. We compare forward KL, reverse KL, and JSD in Table~\ref{tab:ablation_loss}. Reverse KL achieves the best result, while other objectives improve the baselines consistently.

When computing the divergence between the student and the teacher model, we compute all tokens over the vocabulary. Another strategy is to compute probabilities only at the tokens sampled by the student. Table~\ref{tab:ablation_compute} shows that computing probabilities on all tokens performs better, which demonstrates the benefit of richer supervision and teacher knowledge. Other reasons also include that only transferring knowledge of sampled tokens inevitably undermines the mathematical equivalence when we compute KL divergence, and preserving full vocabulary logits yields stable gradient estimates. However, directly transferring logits for a large vocabulary size (usually >100K) becomes expensive. We only transfer last-layer teacher hidden states and recompute logits on the fly with the teacher's head when computing the loss function. Because the dimension of hidden states (e.g., 6144 for Qwen3.5-2B) is much smaller than the dimension of the full vocabulary (e.g., 248320 for Qwen3.5-2B), the corresponding communication overhead becomes affordable and practical, and only introduces negligible recomputation. The infrastructure design is from KDFlow~\citep{zhang2026kdflow}.

In our experiments, we update the teacher fully in sync with the student model, i.e., $\pi^{*}=\pi_{\theta}$ at every training step. We also consider two other variants: (a) fixing the teacher model as the initial weights $\pi^{*}=\pi_{\theta}\vert_{t=0}$ where $t$ denotes training steps, and (b) EMA update: $\pi^{*}\leftarrow m\pi^{*}+\left(1-m\right)\pi_{\theta}$ where $m=0.9$ controls the update rate. We find that fully synchronizing the weights between the teacher and the student achieves the best performance in Table~\ref{tab:ablation_teacher_update}. We hypothesize that the capability of the student model improves during training and can provide better supervision signals.

\begin{figure}[t]
    \centering
    \includegraphics[width=\linewidth]{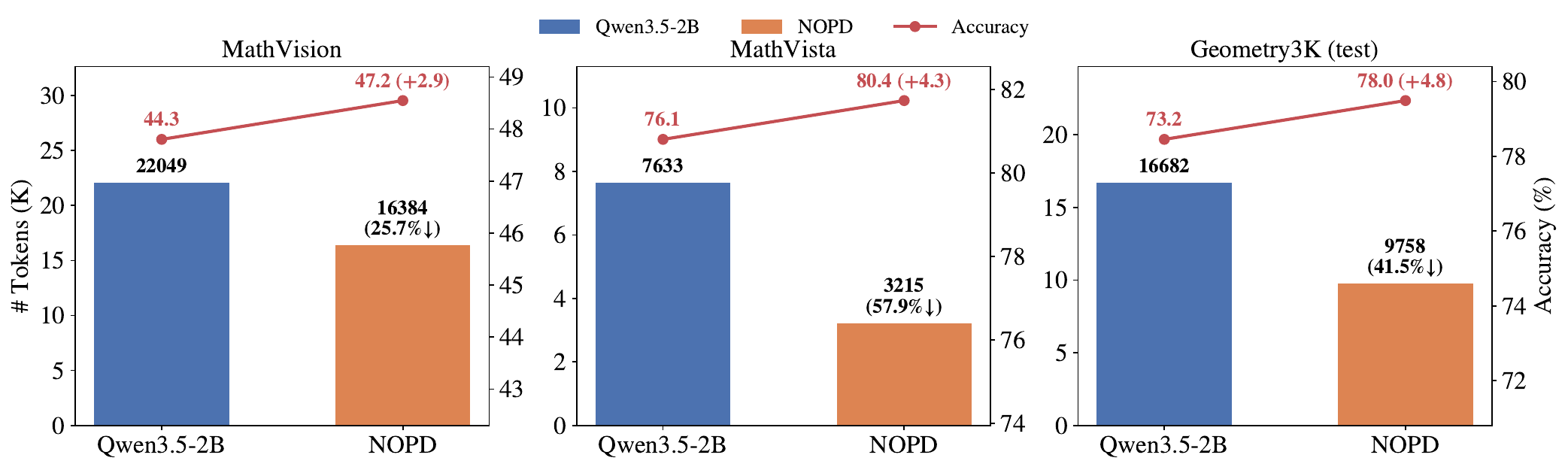}
    \vspace{-1em}
    \caption{The number of output tokens and accuracy on three benchmarks.}
    \vspace{-1em}
    \label{fig:token_acc}
\end{figure}
\begin{figure}[t]
    \centering
    \includegraphics[width=\linewidth]{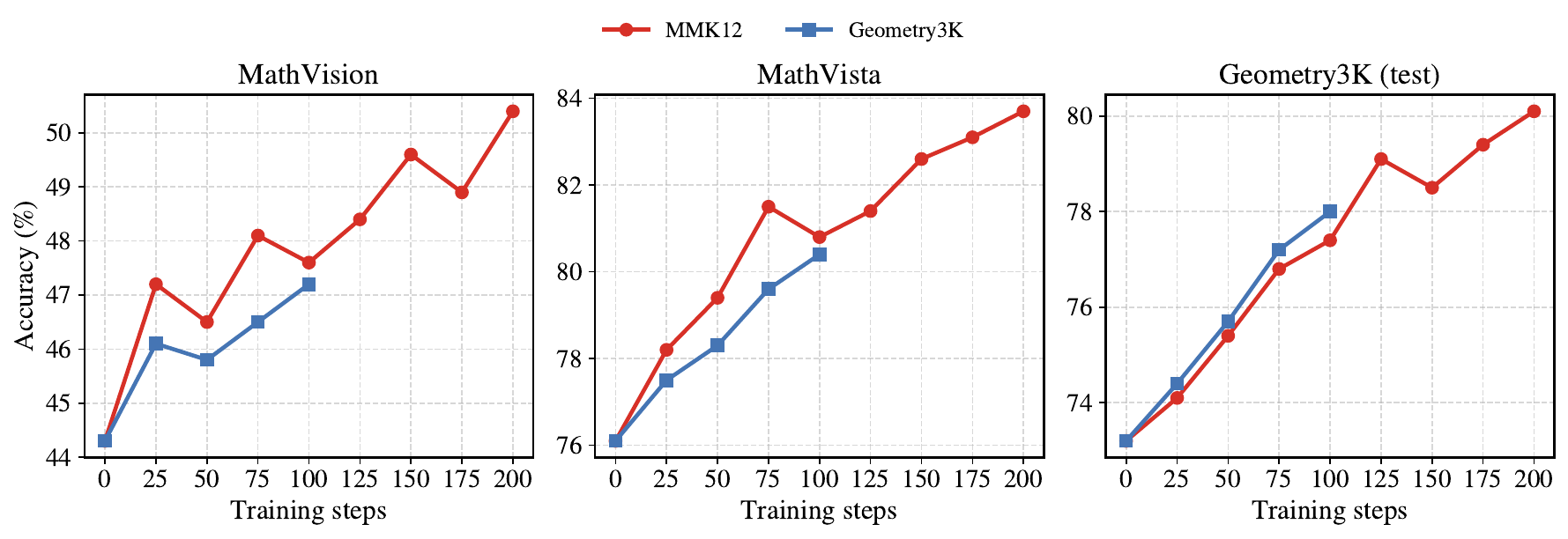}
    \vspace{-1em}
    \caption{\textbf{Accuracy on three benchmarks during training with two different training sets}. We train Qwen3.5-2B with MMK12 and Geometry3K, respectively.}
    \vspace{-1em}
    \label{fig:data_exp}
    \vspace{-1em}
\end{figure}

To reduce computational overhead, we constrain the generation length to 2048 for the student. We explore different generation lengths in Table~\ref{tab:ablation_gen_len}. We find that increasing the generation length to 4096 brings only limited gains compared with other baselines. To balance computation and performance, we set the maximum generation length to 2048. The intuition behind this is that early tokens are more critical for learning in the autoregressive decoding process. Errors in early tokens can accumulate and lead to incorrect outputs. Therefore, providing supervision signals only for early tokens is sufficient for the student's learning.
\vspace{-.5em}
\paragraph{Implicit reasoning compression.}
We compare the number of output tokens and accuracy on three datasets using Qwen3.5-2B in Figure~\ref{fig:token_acc}. Besides the improvement in accuracy, we find that our method simultaneously compresses the thinking process across three benchmarks, which improves the efficiency of reasoning. To be specific, on MathVista, our method reduces tokens by \textbf{57.9\%} and improves accuracy by 4.3\%. Although we do not design a specific technique for compressing tokens, \ours{} achieves implicit reasoning compression, improving efficiency at test time.
\vspace{-.5em}
\paragraph{Effectiveness across training datasets.}
To evaluate the effectiveness of our method across different training datasets, we conduct experiments on MMK12~\citep{mm-eureka}, which contains 15.6K samples. All other settings remain unchanged, except that we increase the training steps to 200 due to the larger dataset size compared to Geometry3K. As shown in Figure~\ref{fig:data_exp}, \ours{} consistently improves performance when trained on MMK12, demonstrating strong scalability with increased data and training compute. For instance, accuracy on MathVision improves by 3.2 points, from 47.2\% to 50.4\%.
\paragraph{Corruption type of images.}
We conduct experiments with different image corruption types to verify their importance. As shown in Table~\ref{tab:ablation_corruption}, all corruptions improve the base model, with adding noise yielding the largest gain ($73.2 \rightarrow 77.1$) and combining all types performing best. This highlights the importance of image corruption for student inputs.

\begin{wraptable}{r}{0.4\textwidth} 
  \centering
  \caption{Ablation on corruption. N, G, L denotes noise, gamma transformation, lower resolution, respectively. We report results on Geometry3K test set with Qwen3.5-2B.}
  \vspace{-.5em}
  \label{tab:ablation_corruption}
  \begin{tabular}{cccc}
    \toprule
    \textbf{N} & \textbf{G} & \textbf{L} & \textbf{Geometry3K} \\
    \midrule
     \yes & \yes & \yes & 78.0 \\
     \yes & & & 77.1\\
      & \yes & &  74.4\\
      & & \yes &  76.8\\
      & & & 73.6 \\
     \midrule
     \multicolumn{3}{c}{\textcolor{lightgray}{w/o training}} & \textcolor{lightgray}{73.2} \\
    \bottomrule
  \end{tabular}
\vspace{-2.5em}
\end{wraptable}
\subsection{Could \ours{} be a General Approach to Improve VLMs?}
\label{sec:exp_general}
\paragraph{Models.}
To validate whether our method can be used to improve VLMs on different tasks and to assess its scalability with data, we conduct experiments using larger-scale datasets and evaluate on multiple tasks. We choose Qwen2.5-VL-7B~\citep{qwen2p5-vl} as the instruction model and Qwen3-VL-Thinking~\citep{bai2025qwen3vl} as the thinking model. To further examine the effectiveness of our method across different models, we include GLM4.6-V-Flash with 9B parameters~\citep{glm45v}. 

\paragraph{Training \& data.}
We collect data from multiple sources, including Geometry3K~\citep{geo3k}, GeomVerse~\citep{kazemi2023geomverse}, Geo170K~\citep{geo170k}, MMK12~\citep{mm-eureka}, GeoQA+~\citep{geoqa_plus}, CoSyn~\citep{yang_CoSyn_2020}, VQAv2~\citep{goyal2017vqav2}, ArxivQA~\citep{li_arxivqa_2024}, ChartQA~\citep{masry2022chartqa} and GQA~\citep{hudson2019gqa}. 

We remove similar images or questions to improve the diversity of the training set. In addition, to ensure vision relevance so that all questions must be answered with images, we prompt Qwen3-VL-8B-Instruct and remove image-irrelevant questions. To balance the training data distribution, we classify all questions into three domains: general, reasoning, and chart, and uniformly sample different domains during training. We get 19K/56K/54K samples for general, chart, and reasoning domains, respectively. We increase training steps by 1000 and keep other hyperparameters the same.
\vspace{-1em}
\paragraph{Evaluation.}
To provide a comprehensive evaluation across diverse tasks, we assess all models on 12 benchmarks covering three categories, including general visual question answering (VQA), reasoning, and chart understanding. Specifically, for general VQA, we adopt MMStar~\citep{mmstar}, MMBench-EN~\citep{liu2024mmbench}, BLINK~\citep{fu2024blink}, CV-Bench~\citep{tong2024cambrian}, and Realworld-QA.\footnote{\url{https://huggingface.co/datasets/xai-org/RealworldQA}} For reasoning tasks, we utilize MathVista~\citep{lu2024mathvista}, MathVision~\citep{mathvision}, MMMU~\citep{mmmu}, and MMMU-Pro~\citep{mmmu-pro}. For chart understanding, we evaluate models on ChartQA-Pro~\citep{masry2025chartqapro}, ChartMuseum~\citep{tang2025chartmuseum}, and CharXiv (reasoning)~\citep{wang2024charxiv}.

For different models, we adopt decoding parameters recommended by the official guide (see Table~\ref{tab:decoding_parameters}) and follow the evaluation protocol in \texttt{lmms-eval}~\citep{lmms-eval}.

\paragraph{Results.}
We present results in Table~\ref{tab:general_exp}. First, it is noticed that \ours{} achieves consistent gains across three models. Specifically, we improve Qwen2.5-VL-7B on \textbf{all} benchmarks. For Qwen3-VL-8B-Thinking and GLM-4.6V-Flash, we get improvements on 10 out of 12 benchmarks. This shows that \ours{} could be used as a general self-improvement approach for VLMs. Furthermore, \ours{} yields substantial improvements across all three domains, which shows the effectiveness of \ours{} is not limited to reasoning tasks but extends to other domains such as VQA and chart understanding.
\newcommand{\plusvalue}[1]{\hspace{0.3em}\textcolor{darkergreen}{(+#1)}}
\newcommand{\minusvalue}[1]{\hspace{0.3em}\textcolor{red2}{(-#1)}}

\definecolor{tblred}{HTML}{FF8988}
\definecolor{tblorange}{HTML}{FECC81}
\definecolor{tblblue}{HTML}{6098FF}
\definecolor{tblgreen}{HTML}{77B25D}
\definecolor{tblpurple}{HTML}{B28CFF}
\definecolor{tblgray}{HTML}{8B8F87}

\begin{table}[t]
\centering
\caption{\textbf{Empirical results on multimodal benchmarks with different models.} \textcolor{darkergreen}{+x} and \textcolor{red2}{-x} indicate improvement/decline compared with initial models.}
\resizebox{\textwidth}{!}{%
    \begin{tabular}{ll|cc|cc|cc}
    \toprule
    & &\textbf{Qwen2.5} & \textbf{\ours} & \textbf{Qwen3-VL} & \textbf{\ours} & \textbf{GLM-4.6V} & \textbf{\ours}  \\
    &  &\textbf{-VL-7B} &  \textbf{-7B} & \textbf{-8B-Thinking} & \textbf{-8B} & \textbf{-Flash} & \textbf{-9B} \\
    \midrule
   
    \cellcolor{tblpurple!20}& MMStar        & 62.5 & 69.9{\tiny\plusvalue{7.4}}  & 73.6 & 75.2{\tiny\plusvalue{1.6}} & 74.3 &  76.8{\tiny\plusvalue{2.5}} \\

    \cellcolor{tblpurple!20} & MMBench-EN    & 83.4 & 83.6{\tiny\plusvalue{0.2}} & 90.6 & 90.4{\tiny\minusvalue{0.2}} & 91.0 & 90.8{\tiny\minusvalue{0.2}} \\

     \cellcolor{tblpurple!20}& BLINK         & 56.4 & 62.3{\tiny\plusvalue{5.9}} & 64.7 & 64.1{\tiny\minusvalue{0.6}} & 65.5  & 66.4{\tiny\plusvalue{0.9}} \\

    \cellcolor{tblpurple!20} & CV-Bench      & 80.0 & 84.3{\tiny\plusvalue{4.3}} & 84.8 &  85.6{\tiny\plusvalue{0.8}} & 86.0 & 86.3{\tiny\plusvalue{0.3}} \\
    \multirow{-5}{*}{\rotatebox{90}{
    \cellcolor{tblpurple!20}
    \small General VQA
}}
& Realworld-QA  & 68.5 & 73.5{\tiny\plusvalue{5.0}}  & 71.9 & 73.4{\tiny\plusvalue{1.5}} & 76.9 & 76.9{\tiny\plusvalue{0.0}} \\
    \midrule
     \cellcolor{tblblue!20} & MathVista & 68.6 & 78.4{\tiny\plusvalue{9.8}} & 78.5  & 81.4{\tiny\plusvalue{2.9}} & 82.9 &  83.3{\tiny\plusvalue{0.4}} \\
     \cellcolor{tblblue!20} & MathVision & 22.4 & 34.4{\tiny\plusvalue{12.0}} & 59.6  & 62.1{\tiny\plusvalue{2.5}} & 54.1 & 57.5{\tiny\plusvalue{3.4}} \\
     \cellcolor{tblblue!20} & MMMU        & 51.3 & 58.1{\tiny\plusvalue{6.8}} & 73.5 & 75.7{\tiny\plusvalue{2.2}} & 71.2 & 74.2{\tiny\plusvalue{3.0}} \\
     \multirow{-4}{*}{\rotatebox{90}{
    \cellcolor{tblblue!20}
    \small Reasoning
}} & MMMU-Pro & 34.6 & 39.6{\tiny\plusvalue{5.0}} & 60.9 & 62.6{\tiny\plusvalue{1.7}} & 59.9 &  63.7{\tiny\plusvalue{3.8}} \\
    \midrule
      \cellcolor{tblorange!20} & ChartQA-Pro  & 43.3 & 44.3{\tiny\plusvalue{1.0}} & 58.4 & 62.9{\tiny\plusvalue{4.5}} & 62.6 & 62.4{\tiny\minusvalue{0.2}} \\
      \cellcolor{tblorange!20} & ChartMuseum   & 26.8 & 31.4{\tiny\plusvalue{4.6}} & 46.7 & 51.6{\tiny\plusvalue{4.9}} & 49.8 & 51.3{\tiny\plusvalue{1.5}} \\
      \multirow{-3}{*}{\rotatebox{90}{
    \cellcolor{tblorange!20}
    \small Chart
}} 
& CharXiv & 42.5 & 47.5{\tiny\plusvalue{5.0}} & 53.0 & 58.4{\tiny\plusvalue{5.4}} & 59.6 & 63.5{\tiny\plusvalue{3.9}} \\
    \bottomrule
    \end{tabular}%
}
\vspace{-1em}
\label{tab:general_exp}
\end{table}
\section{Related Work}
\vspace{-.5em}
\label{sec:related_work}
\paragraph{Self-improvement for foundation models.} 
Previous work has advanced the self-improvement of foundation models by fine-tuning on self-generated data~\citep{zelikman2022star}, automatically constructing preference data~\citep{deng2024enhancing,dong2025selfboosting}, or using LLMs as judges to select responses or label preference data~\citep{yuan2024self_reward,zhang2024restmcts,rest}. Recently, some self-distillation approaches~\citep{self_distil_zero,self_distillation_reasoner}, where a single model plays both teacher and student roles in the on-policy distillation process, have been proposed to improve itself without additional teachers. However, the above methods are designed for language models and require additional feedback from environments~\citep{reinforcement_sd} or ground-truth labels~\citep{self_distillation_reasoner}. Our \ours{} also belongs to self-distillation approaches but does not need additional feedback or answers to improve VLMs.
\vspace{-.5em}
\paragraph{Vision-language models.}
VLMs that combine visual encoders such as CLIP~\citep{clip} with large language models have shown strong ability to understand and reason with both visual and text information inputs~\citep{llava,qwen2p5-vl}. Recently, the development of VLMs has largely been driven by a relentless pursuit of scale in terms of model and data size~\citep{doshi2025gemini,glm45v,bai2025qwen3vl,team2026kimi}. Post-training with high-quality SFT data and large-scale reinforcement learning plays an important role. However, this leads to a high demand for human labels, distillation from frontier models, and the high sampling cost of reinforcement learning. To tackle this, we present \ours, a self-training approach that eliminates the need for external supervision signals.
\vspace{-.5em}
\paragraph{Post-training approaches.}
Post-training that fine-tunes the pre-trained language models aims to make pre-trained models follow human instructions and give useful and helpful responses~\citep{instructgpt}. Typical post-training includes supervised fine-tuning~\citep{wei2022finetuned}, direct preference optimization~\citep{dpo}, reinforcement learning with verifiable rewards~\citep{shao2024deepseekmath,noisyrollout} or from human feedback~\citep{instructgpt}, and distillation from stronger models in an off-policy or on-policy manner~\citep{agarwal2024onpolicy,gu2024minillm,kim2016sequencekd}.
\vspace{-1em}
\section{Conclusion}
We introduce Noisy Student On-Policy Self-Distillation (\ours), a simple yet effective approach to improve VLMs without requiring ground-truth answers or external models. Our key insight is that forcing models to learn in difficult environments, e.g., noisy images, improves their capability across different tasks. Unlike traditional distillation approaches that require teacher models, we use the outputs of the same model under clean inputs as the dense token-level supervision signal. Experimental results show that \ours{} improves the reasoning ability of VLMs across both in-distribution and out-of-distribution sets. Furthermore, we demonstrate that \ours{} can serve as a general approach to improve VLMs across different models and tasks.
\clearpage
\bibliographystyle{plain} 
\bibliography{ref}
\clearpage
\appendix
\section{Limitations and Social Impact}
\label{sec:appendix_limitation}
\subsection{Limitations.}
We propose \ours, an effective self-boosting approach that improves VLMs without external information. The one limitation of our approach is that it is only used in vision-language models because it relies on data augmentations for vision inputs.

\subsection{Social Impact}
\paragraph{Positive impacts.} 
We study post-training and self-improvements for VLMs, which improves the ability of foundation models.

\paragraph{Negative impacts.}
This paper does not inherently involve or encourage applications with direct negative societal impacts. We focus on a general approach to improve VLMs.

\section{Experiment Details}
\subsection{Evaluation Details}
\label{sec:appendix_eval_details}
\paragraph{Decoding \& judge.}
For all models, we use vLLM to accelerate inference and follow the official guide to set the decoding parameters and summarize them in Table~\ref{tab:decoding_parameters}. For Qwen2.5-VL-7B, we report pass@1 accuracy because we use greedy decoding. For other models, we report avg@16 to provide stable results. For tasks that need the language model to extract and judge answers, we use GPT-OSS-120B.\footnote{\url{https://huggingface.co/openai/gpt-oss-120b}}

\paragraph{Benchmark-specific choices.}
In the preliminary experiments of Sec.~\ref{sec:exp_pre_reason}, we evaluate the test/validation split of Geometry3K in an open-ended manner, which differs from the multiple-choice evaluation in \texttt{lmms-eval}. For MMMU and MMMU-Pro with Qwen2.5-VL-7B, we report the maximum performance between the direct answer and the CoT prompt, following the official guide. For ChartMuseum, we report results on the test set. For CharXiv, we report results on the validation set for the reasoning split.

\subsection{Training Details}
We summarize the training hyperparameters in Table~\ref{tab:training_hp}.

\section{More Results and Analysis}
\label{sec:appendix_results_analysis}

\paragraph{Comparison with OPSD.}
OPSD~\citep{self_distillation_reasoner} also belongs to on-policy self-distillation where a single model plays both teacher and student roles. OPSD uses the ground-truth solution as additional privileged information input to the teacher model to provide a high-quality supervision signal. Formally, we reformat it in training VLM scenarios:
\begin{equation}
    \mathcal{L}_{\text{OPSD}}\left(\theta\right) = \mathbb{E}_{\left(q,x\right) \sim \mathcal{D},\, y\sim \pi_{\theta}\left( \cdot \mid q,x\right) } 
       \left[ \mathcal{D}_{\textrm{KL}}\left(\pi_{\theta}\left(y\mid q,x\right) \| \pi_{\theta}\left(y\mid q,x,y^{*}\right)\right)\right].
\end{equation}
Unlike OPSD, our method \ours{} does not need ground-truth solutions to provide additional information. In addition, OPSD and \ours{} are also orthogonal, which means we could combine OPSD and \ours{}. We conduct preliminaries following the section in Sec.~\ref{sec:exp_pre_reason} and results are shown as follows.

\begin{table}[htbp]
\centering
\caption{\textbf{Comparison with OPSD.} }
    \begin{tabular}{lcccccc}
    \toprule
     & \multicolumn{2}{c}{\textbf{Geometry3K}}  & \textbf{MathVista} & \textbf{MathVision}  & \textbf{MMMU}  & \textbf{MMMU-Pro} \\
     & test & val & testmini & test & val & overall \\
    \midrule
    Qwen3.5-2B & 73.2 & 65.3 & 76.1 & 44.3 & 64.2 & 50.3 \\        
    OPSD       & 76.4 & 67.8 & 79.5 & 44.9 & 66.1 & 50.5\\
    \ours      & 78.0 & 69.1 & \textbf{80.4} & 47.2 & 67.6 & 54.7 \\
    OPDS+\ours & \textbf{78.5} & \textbf{70.4} & 80.2 & \textbf{47.5} & \textbf{68.8} & \textbf{55.6}\\
    \bottomrule
    \end{tabular}
\label{tab:compare_opsd}
\end{table}
From results of Table~\ref{tab:compare_opsd}, we can see that \ours{} is more effective than OPSD while eliminating the need for ground-truth solutions. Furthermore, combining OPSD and \ours{} could enhance performance.

\newcommand{\nr}{NoisyRollout}
\paragraph{Difference with \nr.}
\ours{} and \nr~\citep{noisyrollout} share a similar part of adding corruption to images for VLMs. However, the motivations behind this are totally different. \nr{} is a variant of GRPO that adds noise to images to \textit{improve exploration in reinforcement learning}. \nr{} still belongs to RLVR and needs a verifiable answer as we compare in Figure~\ref{fig:compare_approach}. Our \ours{} uses discrepancies between clean and distorted images as a self-supervision signal to improve VLMs without an external model or answers.

\paragraph{Unleashing the potential of \ours{} for open-ended generation tasks.}
We mainly evaluate \ours{} on verifiable tasks, such as reasoning and question answering. Since on-policy distillation provides a \textit{dense} token-level supervision signal, it is natural to ask: is \ours{} effective on open-ended generation tasks, such as image captioning? We conduct preliminary experiments on image captioning tasks. We randomly sample 50K images from ShareGPT-4o\footnote{\url{https://sharegpt4o.github.io/}} and train Qwen2.5-VL-7B for 200 steps.

\begin{figure}[t]
    \centering
    \includegraphics[width=\linewidth]{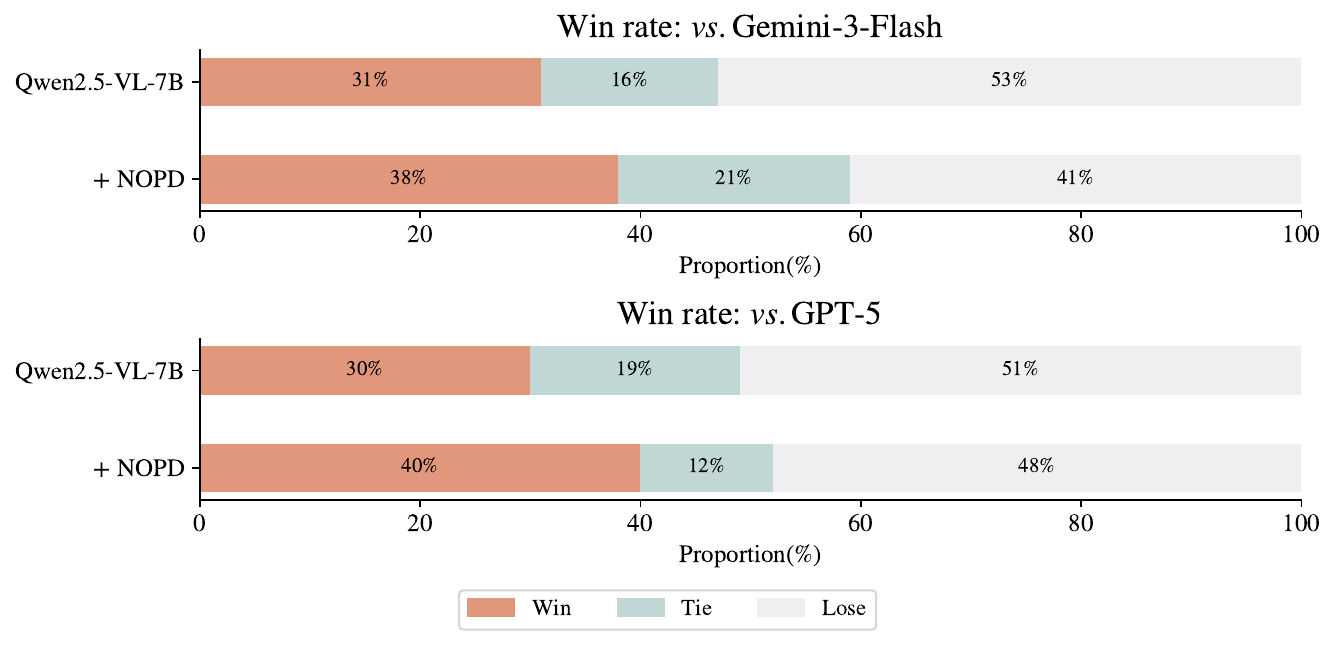}
    \caption{Win-rate comparison on CapArena with Gemini-3-Flash and GPT-5.}
    \label{fig:exp_caption}
\end{figure}

We evaluate the ability of image captioning on CaptionQA~\citep{yang2025captionqa} and CapArena~\citep{cheng2025caparena}. We report accuracy on CaptionQA and win rate against Gemini-3.0-Flash and GPT-5 on CapArena with 100 samples. Results on CaptionQA are shown in Table~\ref{tab:exp_caption}. From Table~\ref{tab:exp_caption}, it can be seen that \ours{} improves Qwen2.5-VL-7B on all domains. Specifically, in natural domains, \ours{} achieves 13-point gains. We also use GPT-4o as a VLM judge against GPT-5 and Gemini-3-Flash in Figure~\ref{fig:exp_caption} on the CapArena Benchmarks. With \ours{}, a model with 7B parameters can match the performance of frontier models. The quality of captions on more than 50\% of images outperforms and matches proprietary models such as Gemini-3-Flash and GPT-5.

\paragraph{Gradient derivation.}
First, we detail the gradient of the \ours{} loss as defined in Eq.~\eqref{eq:noisy_opd}:
\begin{equation}
    \nabla_{\theta}\mathcal{L}_{\textrm{NOPD}}(\theta)=\nabla_{\theta}\sum_{t=1}^{|y|}\sum_{v=1}^{|V|} \pi_{\theta}(y_{v,t}|q,\mathcal{T}(x)) \log \frac{\pi_{\theta}(y_{v,t}|q,\mathcal{T}(x))}{\pi_{\theta}(y_{v,t}|q,x)}.
\end{equation}
Since we stop the gradient of $\pi_{\theta}(y_{v,t}|q,x)$, we denote $\pi_{\textrm{ref}}(y_{v,t}|q,x)=\pi_{\theta}(y_{v,t}|q,x)$ for clarity and ignore the subscript NOPD.

\begin{align}
\nabla_{\theta}\mathcal{L}(\theta) 
&= \nabla_{\theta} \sum_{t=1}^{\lvert y\rvert}\sum_{v=1}^{\lvert V\rvert} 
\pi_{\theta}(y_{v,t}|q,\mathcal{T}(x)) 
\log \frac{\pi_{\theta}(y_{v,t}|q,\mathcal{T}(x))}{\pi_{\textrm{ref}}(y_{v,t}|q,x)} \notag\\
&= \sum_{t=1}^{|y|}\sum_{v=1}^{|V|} 
\log \frac{\pi_{\theta}(y_{v,t}|q,\mathcal{T}(x))}{\pi_{\textrm{ref}}(y_{v,t}|q,x)}
\nabla_{\theta}\pi_{\theta}(y_{v,t}|q,\mathcal{T}(x)) \notag
+ \sum_{t=1}^{|y|}\sum_{v=1}^{|V|} 
\pi_{\theta}(y_{v,t}|q,\mathcal{T}(x))
\nabla_{\theta}\log \frac{\pi_{\theta}(y_{v,t}|q,\mathcal{T}(x))}{\pi_{\textrm{ref}}(y_{v,t}|q,x)} \\
&= \sum_{t=1}^{|y|}\sum_{v=1}^{|V|} 
\log \frac{\pi_{\theta}(y_{v,t}|q,\mathcal{T}(x))}{\pi_{\textrm{ref}}(y_{v,t}|q,x)}
\nabla_{\theta}\pi_{\theta}(y_{v,t}|q,\mathcal{T}(x)) \notag
+ \sum_{t=1}^{|y|}\sum_{v=1}^{|V|} 
\pi_{\theta}(y_{v,t}|q,\mathcal{T}(x)) \frac{\nabla_{\theta}\pi_{\theta}(y_{v,t}|q,\mathcal{T}(x))}{\pi_{\theta}(y_{v,t}|q,\mathcal{T}(x))} \\
&= \sum_{t=1}^{|y|}\sum_{v=1}^{|V|} 
\log \frac{\pi_{\theta}(y_{v,t}|q,\mathcal{T}(x))}{\pi_{\textrm{ref}}(y_{v,t}|q,x)}
\nabla_{\theta}\pi_{\theta}(y_{v,t}|q,\mathcal{T}(x)) \notag
+ \sum_{t=1}^{|y|}\sum_{v=1}^{|V|} \nabla_{\theta}\pi_{\theta}(y_{v,t}|q,\mathcal{T}(x)) \\
&= \sum_{t=1}^{|y|}\sum_{v=1}^{|V|} 
\log \frac{\pi_{\theta}(y_{v,t}|q,\mathcal{T}(x))}{\pi_{\textrm{ref}}(y_{v,t}|q,x)}
\nabla_{\theta}\pi_{\theta}(y_{v,t}|q,\mathcal{T}(x)) \notag
+ \sum_{t=1}^{|y|}\nabla_{\theta}\sum_{v=1}^{|V|} \underbrace{\pi_{\theta}(y_{v,t} \mid q, \mathcal{T}(x))}_{1} \\
&= \sum_{t=1}^{|y|}\sum_{v=1}^{|V|} 
\log \frac{\pi_{\theta}(y_{v,t}|q,\mathcal{T}(x))}{\pi_{\textrm{ref}}(y_{v,t}|q,x)}
\nabla_{\theta}\pi_{\theta}(y_{v,t}|q,\mathcal{T}(x)) \notag\\
&= \sum_{t=1}^{|y|}\sum_{v=1}^{|V|} 
\log \frac{\pi_{\theta}(y_{v,t}|q,\mathcal{T}(x))}{\pi_{\textrm{ref}}(y_{v,t}|q,x)} \pi_{\theta}(y_{v,t}|q,\mathcal{T}(x))  \nabla_{\theta}\log \pi_{\theta}(y_{v,t}|q,\mathcal{T}(x)) \notag\\
&=\sum_{t=1}^{|y|} 
\mathbb{E}_{\hat{y}_t \sim \pi_\theta(\cdot \mid \mathcal{T}(x), q, y_{<t})}
\quad \Bigg[\log \frac{\pi_\theta(\hat{y}_t \mid q, \mathcal{T}(x))}
         {\pi_\theta(\hat{y}_t \mid q, x)} 
\cdot \nabla_\theta \log \pi_\theta(\hat{y}_t \mid q, \mathcal{T}(x), y_{<t}) \Bigg].
\end{align}
Finally, we get the gradient of $\mathcal{L_{\textrm{NOPD}}}$ in Eq.~\eqref{eq:grad_nopd}.
\begin{table}[t]
\centering
\caption{Summary of decoding parameters during evaluation.}
\resizebox{\textwidth}{!}{%
    \begin{tabular}{lcccccc}
    \toprule
    \textbf{} & \textbf{} & \textbf{} & \textbf{} & \textbf{Presence} & \textbf{Repetition} & \textbf{} \\
    \textbf{Model} & \textbf{Temperature} & \textbf{Top-p} & \textbf{Top-k} & \textbf{penalty} & \textbf{penalty} & \textbf{Max tokens}\\
    \midrule
    Qwen2.5-VL-7B & 0 & 1.0 & 1 & 1.0 & 1.0 & 2048 \\
    Qwen3.5-2B    & 1.0 & 0.95 & 20 & 1.5 & 1.0 & 32768 \\
    Qwen3-VL-8B-Thinking  & 1.0 & 0.95 & 20 & 0.0 & 1.0 & 40960\\
    GLM-4.6V-Flash & 0.8 & 0.6 & 2 & 1.0 & 1.1 & 16384 \\
    \bottomrule
    \end{tabular}%
}
\label{tab:decoding_parameters}
\end{table}
\begin{table}[t]
  \centering
  \caption{Training hyperparameters. ``X/Y'' denotes that we use X for Qwen2.5-VL-7B and Y for Qwen3.5-2B.}
  \label{tab:training_hp}
  \begin{tabular}{lccccc}
  \toprule
               & SFT & Off-polocy KD & RLVR & OPD & \ours \\
    \midrule
    Training backend & \multicolumn{5}{c}{FSDP2} \\
    
    Precision & \multicolumn{5}{c}{BF16} \\
    Gradient clip & \multicolumn{5}{c}{1.0} \\
    Attention backend & \multicolumn{5}{c}{FlashAttention 2} \\ 
    Optimizer &   \multicolumn{5}{c}{AdamW} \\
    Learning rate & \multicolumn{5}{c}{1e-6} \\
    Inference engine & -& - & \multicolumn{3}{c}{SGLang} \\
    LRScheduler & cosine & cosine & constant & constant & constant\\    
    Epoch / iter & 3 epoch & 3 epoch & 100 steps & 100 steps & 100 steps \\
    Global batch size & 64 & 64 & 128 & 32 & 32\\    
    Rollout temperature & - & - & 1.0 & 1.0 & 1.0\\
    Rollout top-p & - & - & 0.95 & 0.95 & 0.95 \\
    Rollout number & - & - & 8 & 1 & 1 \\
    Max gen length & - & - & 2048/16384 & 2048 & 2048 \\
    \bottomrule    
  \end{tabular}
\end{table}
\paragraph{Does \ours{} work by improving diversity?}
We found that injecting vision permutation $\mathcal{T}$ slightly changes the distribution of student models. This naturally raises the question: does the improvement of \ours{} come from the diversity of rollouts? To disentangle the effects of diversity and vision permutation, we conduct experiments under different rollout temperatures. For each question, we sample 5 responses and score their diversity using an LLM as a judge on a scale from 1 to 10. Note that we keep the number of rollout response to 1 when computing gradients to align other settings. Other questions are only used to examine diversity. 

Results are shown in Table~\ref{tab:exp_temp_diversity}. We found that with image transformation $\mathcal{T}$, the diversity of rollout outputs does not improve at temperatures of 0.8 and 1.0, yet performance increases. At a temperature of 1.2, diversity continues to improve but performance drops. These results indicate that the performance gains of \ours{} do not stem from higher diversity.

\begin{table}[t]
\centering
\caption{Results on CaptionQA.}
\begin{tabular}{lccccc}
    \toprule
    \textbf{Model} & \textbf{Natural} & \textbf{Document} & \textbf{E-comm} & \textbf{Embodied} & \textbf{Avg}. \\
    \midrule
    
    GPT5 & 88.15 & 89.98 & 95.44 & 85.30 & 89.72 \\
    Gemini-3-Flash & 87.66 & 84.59 & 89.97 & 84.03 & 86.56 \\
    Gemini-3.1-Pro & 88.49 & 90.82 & 90.36 & 85.15 & 88.71 \\
    Qwen2.5-VL-7B & 71.64 & 75.85 & 85.38 & 68.36 & 75.31 \\
    + \ours & 84.68 & 82.70 & 87.64 & 74.19 & 82.31 \\
    \bottomrule
    \end{tabular}
\label{tab:exp_caption}
\end{table}

\paragraph{Visualization examples.} We provide two example visualizations in Table~\ref{tab:example_1} and Table~\ref{tab:example_2} to demonstrate that \ours{} improves the perception and reasoning abilities compared with the base model. In particular, Qwen2.5-VL-7B fails to properly analyze the relative relationships between the line segments in the figure (Table~\ref{tab:example_1}) and to correctly identify which segment the labeled length refers to (Table~\ref{tab:example_2}).
\begin{table}[t]
\centering
\caption{Ablation on different rollout temperature.}    
    \begin{tabular}{lccc}
    \toprule
     & Temperature & Accuracy & Diversity \\
     \midrule
     w/o $\mathcal{T}$ & 0.8 & 73.1 & 7.2\\
                       & 1.0 & 73.6 & 7.4\\
                       & 1.2 & 73.0 & 7.5\\
    \midrule
     w/ $\mathcal{T}$  &  0.8 & 78.0 & 7.1\\
                       &  1.0 & 78.0 & 7.4\\
                       &  1.2 & 77.8 & 7.8\\
    
    \bottomrule
    \end{tabular}
\label{tab:exp_temp_diversity}
\end{table}
\begin{table}[t]
\centering
\caption{Case study of a visualized example from MathVision with Qwen2.5-VL-7B. We \textcolor{red}{highlight errors} generated by Qwen2.5-VL-7B.}
\label{tab:example_1}
\begin{tabular}{p{0.9\linewidth}}
\toprule
\includegraphics[width=0.18\linewidth]{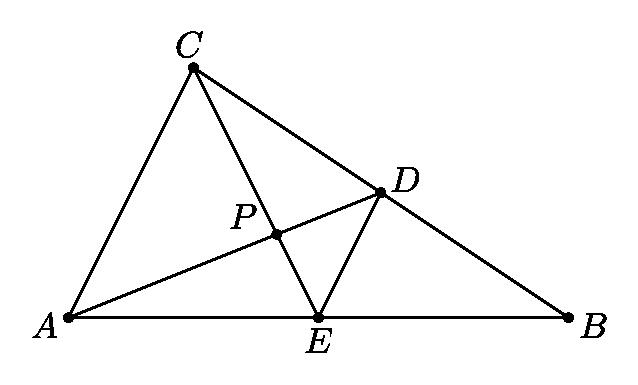}
Question: In $\triangle ABC$, medians $\overline{AD}$ and $\overline{CE}$ intersect at $P$, $PE=1.5$, $PD=2$, and $DE=2.5$. What is the area of $AEDC$? Answer: 13.5 \\
\midrule
\rowcolor{gray!20}
\textit{\textbf{Qwen2.5-VL-7B}} \\
To solve the problem, we need to determine the area of quadrilateral \(AEDC\) in \(\triangle ABC\) given the lengths of the segments formed by the intersection of the medians at point \(P\). 1. **Identify the properties of the medians and centroid:** - The medians of a triangle intersect at the centroid \(P\), which divides each median into a ratio of 2:1. - Therefore, \(AP:PD = 2:1\) and \(CP:PE = 2:1\). 2. **Determine the lengths of the segments:** - Given \(PE = 1.5\) and \(PD = 2\), we can find \(PD\) and \(PE\) in terms of the total length of the median. - Since \(PD = 2\) and \(PE = 1.5\), \textcolor{red}{the total length of the median \(CE\) is \(PD + PE = 2 + 1.5 = 3.5\).} - The centroid \(P\) divides \(CE\) into segments \(CP\) and \(PE\) in the ratio 2:1. Therefore, \(CP = 2 \times 1.5 = 3\) and \(PE = 1.5\). 3. **Calculate the area of \(\triangle AEDC\):** - The area of \(\triangle AEDC\) can be found by considering the areas of the smaller triangles formed by the medians. - \textcolor{red}{The centroid divides the triangle into six smaller triangles of equal area. - The area of \(\triangle AEDC\) is the sum of the areas of \(\triangle APE\) and \(\triangle CPE\)}. 4. **Use the given lengths to find the area:** - \textcolor{red}{The area of \(\triangle AEDC\) is the sum of the areas of \(\triangle APE\) and \(\triangle CPE\)}. - The area of \(\triangle APE\) is \(\frac{1}{3}\) of the area of \(\triangle AEC\) and the area of \(\triangle CPE\) is \(\frac{1}{3}\) of the area of \(\triangle CED\). - The area of \(\triangle AEDC\) is the sum of the areas of \(\triangle APE\) and \(\triangle CPE\), which is \(\frac{2}{3}\) of the area of \(\triangle AEC\). 5. **Calculate the area of \(\triangle AEDC\):** - The area of \(\triangle AEDC\) is \(\frac{2}{3}\) of the area of \(\triangle AEC\). - The area of \(\triangle AEC\) is the sum of the areas of \(\triangle APE\) and \(\triangle CPE\), which is \(\frac{2}{3}\) of the area of \(\triangle AEDC\). 6. **Final answer:** - The area of \(\triangle AEDC\) is \(\boxed{10}\). \no \\
\midrule
\rowcolor{gray!20}
\textit{\textbf{NOPD}} \\
To solve the problem, we aim to find the area of quadrilateral $AEDC$ in $\triangle ABC$ based on the segment lengths determined by the intersection of the medians at point $P$.

1. \textbf{Identify the properties of the medians and centroid:}  
- The medians of a triangle meet at the centroid $P$, which divides each median in a ratio of 2:1 (measured from the vertex to the midpoint).  
- Hence, $AP:PD = 2:1$ and $CP:PE = 2:1$.

2. \textbf{Determine the lengths of the segments:}  
- Given $PD = 2$, and using $PD = \tfrac{1}{3}AD$, we obtain $AD = 3 \times 2 = 6$.  
- Given $PE = 1.5$, and using $PE = \tfrac{1}{3}CE$, we obtain $CE = 3 \times 1.5 = 4.5$.  
- In addition, $DE = 2.5$. Since $D$ and $E$ are midpoints, $DE$ is a midsegment, so $DE = \tfrac{1}{2}AC$, which implies $AC = 5$.

3. \textbf{Relate the medians to the area of the triangle:}  
- The three medians of a triangle form a new triangle whose area is $\tfrac{3}{4}$ of the area of the original triangle.  
- In this case, the medians have lengths $6$, $4.5$, and the third median can be found to be $7.5$.

4. \textbf{Compute the area using the given lengths:}  
- Applying Heron's formula to the triangle formed by the medians:
$
s = \frac{6 + 4.5 + 7.5}{2} = 9
$
$
\text{Area} = \sqrt{9(9-6)(9-4.5)(9-7.5)} = 13.5
$
- Therefore, the area of $\triangle ABC$ is:
$
[ABC] = \frac{4}{3} \times 13.5 = 18
$

5. \textbf{Calculate the area of $AEDC$:}  
- Triangle $BED$ is formed by connecting midpoints, so its area is $\tfrac{1}{4}$ of the area of $\triangle ABC$:
$
[BED] = \frac{1}{4} \times 18 = 4.5
$
- Thus, the area of quadrilateral $AEDC$ is:
$
[AEDC] = [ABC] - [BED] = 18 - 4.5 = 13.5
$

6. \textbf{Final answer:}  
- The area of $\triangle AEDC$ is $\boxed{13.5}$. \yes \\
\bottomrule
\end{tabular}
\end{table}
\begin{table}[t]
\centering
\caption{Case study of a visualized example from from validation set of Geometry3K with Qwen2.5-VL-7B. We \textcolor{red}{highlight errors} generated by Qwen2.5-VL-7B.}
\label{tab:example_2}
\begin{tabular}{p{0.9\linewidth}}
\toprule
\includegraphics[width=0.18\linewidth]{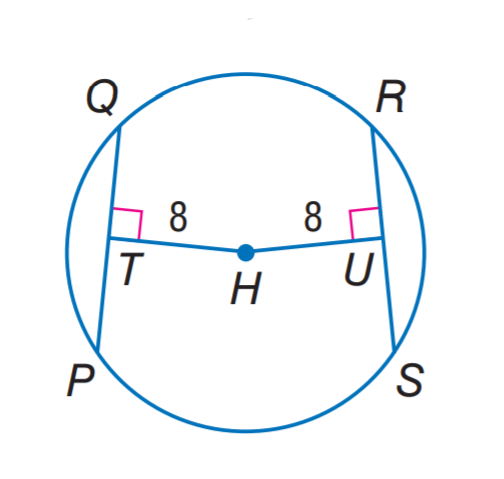}
Question: In $\odot H$, $PQ=3x-4$ and $RS=14$. Find $x$. Answer: 6 \\
\midrule
\rowcolor{gray!20}
\textit{\textbf{Qwen2.5-VL-7B}} \\
 To solve for \( x \) in the given problem, we need to use the properties of circles and chords. Here's the step-by-step reasoning: 1. **Identify the given information:** - The circle is centered at \( H \). - \( PQ = 3x - 4 \). - \( RS = 14 \). 2. **Understand the relationship between the chords and the diameter:** - Since \( T \) and \( U \) are points on the circle such that \( TH \perp PQ \) and \( UH \perp RS \), these perpendiculars bisect the chords \( PQ \) and \( RS \) respectively. - Therefore, \( PT = TQ = \frac{PQ}{2} \) and \( RU = US = \frac{RS}{2} \). 3. **Calculate the lengths of the segments:** - Given \( PQ = 3x - 4 \), the length of \( PT \) and \( TQ \) is: \[ PT = TQ = \frac{PQ}{2} = \frac{3x - 4}{2} \] - Given \( RS = 14 \), the length of \( RU \) and \( US \) is: \[ RU = US = \frac{RS}{2} = \frac{14}{2} = 7 \] 4. **Set up the equation using the total length of the chord \( PQ \):** - Since \( PQ = PT + TQ \): \[ PQ = \frac{3x - 4}{2} + \frac{3x - 4}{2} = 3x - 4 \] 5. **Solve for \( x \):** - We know \( PQ = 3x - 4 \) and \textcolor{red}{from the diagram, \( PQ = 16 \)} (since \( T \) and \( U \) are midpoints and the total length is twice the segment from the center to the midpoint of the chord). - Set up the equation: \[ 3x - 4 = 16 \] - Solve for \( x \): \[ 3x = 20 \] \[ x = \frac{20}{3} \] Therefore, the value of \( x \) is \(\boxed{\frac{20}{3}}\).\no \\
\midrule
\rowcolor{gray!20}
\textit{\textbf{NOPD}} \\
To solve for \( x \) in the given problem, we need to use the properties of circles and chords. Here's the step-by-step reasoning:

1. \textbf{Identify the given information:}
- The circle is centered at \( H \).
- \( PQ = 3x - 4 \).
- \( RS = 14 \).
- The perpendicular distances from the center to both chords are equal: \( HT = HU = 8 \).

2. \textbf{Understand the relationship between the chords and the center:}
- Since \( HT \perp PQ \) and \( HU \perp RS \), these perpendiculars bisect the chords \( PQ \) and \( RS \) respectively.
- Therefore, \( PT = TQ = \frac{PQ}{2} \) and \( RU = US = \frac{RS}{2} \).

3. \textbf{Use the equal distance property of chords:}
- In a circle, chords that are equidistant from the center are equal in length.
- Since \( HT = HU = 8 \), chords \( PQ \) and \( RS \) are equal.
- Hence, \( PQ = RS \).

4. \textbf{Set up the equation:}
\[
3x - 4 = 14
\]

5. \textbf{Solve for \( x \):}
\[
3x = 18
\]
\[
x = 6
\]

6. \textbf{Final answer:}
- The value of \( x \) is \(\boxed{6}\). \yes \\
\bottomrule
\end{tabular}
\end{table}
\clearpage
\end{document}